\documentclass[twoside]{article}

\usepackage[accepted]{aistats2025}

\usepackage[T1]{fontenc}
\usepackage[utf8]{inputenc}
\usepackage{amsthm}
\usepackage{amsmath}
\usepackage{amssymb}
\usepackage{graphicx}
\usepackage{dsfont}
\usepackage{amsmath}
\usepackage{array}
\usepackage{multirow}
\usepackage{caption}
\usepackage{color}
\usepackage{ifthen}

\usepackage{multicol}

\theoremstyle{plain}

\theoremstyle{plain}

\theoremstyle{plain}
\newtheorem{lem}{\protect\lemmaname}
\theoremstyle{plain}
\newtheorem{thm}{\protect\theoremname}
\theoremstyle{plain}
  
\theoremstyle{plain}

\theoremstyle{definition}

\theoremstyle{definition}
\newtheorem{assump}{\protect\assumptionname}
\theoremstyle{definition}

\makeatother
  
\usepackage{babel} 

\providecommand{\claimname}{Claim}
\providecommand{\lemmaname}{Lemma}
\providecommand{\propositionname}{Proposition}
\providecommand{\theoremname}{Theorem}
\providecommand{\corollaryname}{Corollary} 
\providecommand{\factname}{Fact} 
\providecommand{\definitionname}{Definition}
\providecommand{\assumptionname}{Assumption}
\providecommand{\remarkname}{Remark}

\DeclareMathOperator*{\argmax}{arg\,max}

\newcommand{\openone}{\mathds{1}}

\newboolean{showedits}
\setboolean{showedits}{true} % toggle to show or hide edits
\ifthenelse{\boolean{showedits}}
{
 \newcommand{\del}[1]{\textcolor{red}{\sout{#1}}} % please delete
}{
 \newcommand{\del}[1]{} % please delete
 
}

\newboolean{showcomments}
\setboolean{showcomments}{true} % ATTN: THIS TURNS ON THE ANNOTATIONS
\ifthenelse{\boolean{showcomments}}
{\newcommand{\nbc}[3]{
 {\colorbox{#3}{\bfseries\sffamily\scriptsize\textcolor{white}{#1}}}
 {\textcolor{#3}{\sf\small$\blacktriangleright$\textit{#2}$\blacktriangleleft$}}}
 }
{\newcommand{\nbc}[3]{}
 \renewcommand{\del}[1]{} % please delete
 }

\definecolor{tdcolor}{rgb}{1.0,0,0}
\newcommand{\kSE}{k_{\text{SE}}}

\newcommand{\GP}{{\rm GP}}

\newcommand{\btheta}{\boldsymbol{\theta}}

\newcommand{\ftil}{\widetilde{f}}

\newcommand{\Nc}{\mathcal{N}}

\newcommand{\xv}{\mathbf{x}}

\newcommand{\yv}{\mathbf{y}}

\newcommand{\Fc}{\mathcal{F}}

\newcommand{\EE}{\mathbb{E}}

\newcommand{\RR}{\mathbb{R}}

\newcommand{\Kv}{\mathbf{K}}
\newcommand{\kv}{\mathbf{k}}

\newcommand{\bzero}{\boldsymbol{0}}

\newcommand{\deldag}{\Delta(\dagger)}

\newcommand{\Otil}{\widetilde{O}}

\usepackage{color}

\newcommand{\mywidth}{0.85\textwidth}
\newcommand{\rev}[1]{{#1}}

\begin{document} 
	
	\twocolumn[
	
	\aistatstitle{Lower Bounds for Time-Varying Kernelized Bandits}
	
	\aistatsauthor{ Xu Cai \And Jonathan Scarlett }
	
	\aistatsaddress{ National University of Singapore \And National University of Singapore } ]

\begin{abstract}
     The optimization of black-box functions with noisy observations is a fundamental problem with widespread applications, and has been widely studied under the assumption that the function lies in a reproducing kernel Hilbert space (RKHS).  This problem has been studied extensively in the stationary setting, and near-optimal regret bounds are known via developments in both upper and lower bounds.  In this paper, we consider non-stationary scenarios, which are crucial for certain applications but are currently less well-understood.  Specifically, we provide the first algorithm-independent lower bounds, where the time variations are subject satisfying a total variation budget according to some function norm.  Under $\ell_{\infty}$-norm variations, our bounds are found to be close to an existing upper bound (Hong \emph{et al.}, 2023).  Under RKHS norm variations, the upper and lower bounds are still reasonably close but with more of a gap, raising the interesting open question of whether non-minor improvements in the upper bound are possible.
\end{abstract}
% \documentclass[english]{article}
% \usepackage[a4paper, total={6in, 8.5in}]{geometry}
% \input{preamble.tex}

% \usepackage{setspace}
% \usepackage{hyperref}
% \usepackage{algorithm}
% \usepackage{algorithmic}
% \usepackage{natbib}
% % \setcitestyle{authoryear,round,citesep={;},aysep={,},yysep={;}}

% \setlength{\parindent}{2em}
% \setlength{\parskip}{0.5em}
% \setstretch{1.3}

% \hypersetup{
%  colorlinks=true,
%  linkcolor=blue,
%  filecolor=magenta, 
%  urlcolor=cyan,
%  citecolor=magenta
% }

% \title{On Optimal Order of Time Varying Kernelized Bandits}
% % \author{Cai Xu}

% \date{}

% \begin{document}

% \maketitle

\section{Introduction}

A fundamental problem in optimization is to optimize a function $f(\xv)$ given only noisy black-box queries.  Kernel-based methods have proved to be highly effective for this task, with theoretical results assuming that the function lies in a {\em reproducing kernel Hilbert space} (RKHS) for some kernel $k$ that dictates the smoothness properties of the function.  This is known as \emph{kernelized bandits}, and also falls under the scope of Bayesian optimization (BO) using Gaussian processes (GP) \cite{garnett2023bayesian}.  Previous works have predominantly focused on stationary $f$, leading to nearly optimal theoretical guarantees (see Section \ref{sec:relate} for a summary).

% Kernelized bandits, also known as Bayesian optimization (BO), or Gaussian process (GP) bandits, has gained popularity in optimizing the value $f(\xv^*)$ of a kernel black-box function $f$, reproduced from kernels $k$ that satisfying reproducing properties.  The kernel $k$ forms a {\em reproducing kernel Hilbert space} (RKHS), that reproduces every function in the space in the sense that the function evaluation at any $\xv$ can be performed by taking an inner product with the kernel, i.e., $f(\xv) = \langle f, k(\cdot,\xv)\rangle_k$.  

Real-world environments often exhibit dynamic changes, sparking recent research interest in \emph{non-stationary} (or \emph{time-varying}) kernelized bandits.  In this context, the objective function $f_t(\xv)$ can vary with the time index $t$.  To obtain meaningful optimization guarantees, it is customary to place limitations on how much the function can very; our focus is on restrictions of the following form:
\begin{equation}
    \sum_{t=1}^{T-1} \|f_{t+1}-f_{t}\|_{\dagger} \le \deldag, \label{eq:restriction}
\end{equation}
where $\{f_1,\ldots,f_T\}$ are $T$ black-box instances at each timestamp, $\dagger\in\{\infty,k\}$ indicates the choice of norm, and the total variation $\deldag$ is some positive value.  (When clear from the context, we simply write $\Delta$, omitting the argument.)  We are primarily interested in the regime $\Delta = o(T)$, meaning that the amount of variation is low compared to the time horizon.

Another type of restriction uses $L\ge 1$ to limit the total {\em number} of changes, namely,
\begin{equation}
	 \sum_{t=1}^T \openone \{f_{t+1}\neq f_t \} \le L-1, \label{eq:L_changes}
\end{equation}
which is suited to handling rare but possibly abrupt changes.  We will briefly consider this setting as well, but will focus mainly on \eqref{eq:restriction} which turns out to be more challenging/interesting when it comes to deriving lower bounds.

We briefly mention some other aspects of our problem setup as follows, deferring the full details to Section \ref{sec:setup}: The function domain is $[0,1]^d$, each function has bounded RKHS norm $\|f_t\|_k \le B$, the black-box queries are corrupted by $\Nc(0,\sigma^2)$ additive noise, and we treat $B$ and $\sigma^2$ as constants that may be hidden in $O(\cdot)$ notation.  We measure performance using the (non-stationary) cumulative regret $R_T = \sum_{t=1}^{T} \big( f_t(\xv^*_t) - f_t(\xv_t) \big)$, where $\xv^*_t$ maximizes $f_t$.

% \medskip
{\bf\noindent Existing Upper Bounds.} The norm choices $\dagger=\infty$ and $\dagger=k$ have both been considered in prior works, seemingly without much consideration for how the two compare.  The two can immediately be related via $|f(\xv)|=\langle f, k(\cdot,\xv)\rangle_k \le \|f(\xv)\|_k \cdot\|k(\cdot,\xv)\|_k$, so under the standard normalization $k(\xv,\xv) = 1$, an RKHS norm variation upper bound of $\Delta$ implies an $\ell_{\infty}$-norm variation upper bound of $\Delta$ (but not vice versa).  Hence, the RKHS norm version is a \emph{potentially easier problem} that could have a smaller cumulative regret.

\rev{For both $\dagger \in \{\infty,k\}$, the previous state-of-the-art upper bound in the kernelized setting (before a concurrent paper to ours which we discuss shortly)} is due to \cite{Hon23}, and is given by\footnote{We adopt the standard notation $\Otil(f_n) = O(f_n {\rm poly}(\log f_n))$, i.e., suppressing log factors.}
\begin{equation}
	R_T \le \Otil\Big( \gamma_T^{\frac{1}{3}} \Delta^{\frac{1}{3}} T^{\frac{2}{3}}\Big),
\end{equation}
where $\gamma_T$ is the \emph{maximum information gain} associated with the kernel (e.g., see \cite{Sri09,Vak21a}, or see Appendix \ref{sec:master_gen} for a formal definition).  
In particular, for the widely-adopted Mat\'ern kernel with smoothness parameter $\nu$, substituting $\gamma_T = \Otil(T^{\frac{\nu}{2\nu + d}})$ \cite{Vak21a} leads to
\begin{equation}
	R_T \le \Otil\big(\Delta^{\frac{1}{3}} T^{\frac{4\nu+3d}{6\nu+3d}}\big) = \Otil\big(\Delta^{\frac{1}{3}} T^{\frac{4\nu/3+d}{2\nu+d}}\big), \label{eq:RT_Matern}
\end{equation}
and in the limit as $\nu \to \infty$, this becomes increasingly close to dependence $\Delta^{1/3}T^{2/3}$, matching the dependence attained for linear bandits \cite{Che19,Wei21}.  See Section \ref{sec:relate} for further discussion on linear bandits.

In the case of at most $L$ switches (see \eqref{eq:L_changes}), the upper bound in \cite{Hon23} is $\Otil( \sqrt{L T \gamma_T} )$, which is $\sqrt{L}$ times the well-known $\Otil( \sqrt{L T \gamma_T} )$ for the stationary setting.  For the Mat\'ern kernel with parameter $\nu$, this becomes
\begin{equation}
	R_T = \Otil\big( \sqrt{L} T^{\frac{\nu+d}{2\nu+d}}\big), \label{eq:RT_Matern_L}
\end{equation}
which becomes increasingly close to dependence $\sqrt{LT}$ as $\nu \to \infty$.

We note that if $\Delta = T^{a}$ for some $a \in (0,1)$, then there exist certain choices of $(a,\nu,d)$ for which \eqref{eq:RT_Matern} fails to be sublinear in $T$, and similarly for \eqref{eq:RT_Matern_L} when $L = T^a$.  In such cases, we can ``improve'' the bound to $O(T)$, and we will do so for the purpose of checking the tightness of our lower bounds.

% Regarding regret bounds, near-optimal bounds have been achieved for linear bandits \cite{Wei21} against the lower bound studied in \cite{Che19}.  However, to our knowledge, there are no existing lower bounds in the kernelized time varying setting.  Therefore, it remains unknown whether the current best kernelized result $\Otil( \gamma_T^{\frac{1}{3}} \Delta^{\frac{1}{3}} T^{\frac{2}{3}})$ from \cite{Hon23} can be improved. 

% However, a critical limitation lurks when we substitute the upper bound of $\gamma_T$ \cite{Vak21a} for the Mat\'ern kernel and reduce it to $\Otil(\Delta^{\frac{1}{3}} T^{\frac{4\nu+3d}{6\nu+3d}})$: when $\Delta$ varies with $T$ exceeding a specific power term (e.g., $\frac{6\nu}{6\nu+3d}$, yet $\Delta$ still remaining sublinear), the overall bound loses its sublinearity in $T$.  This scenario, unfortunately, arises often in practice, particularly with highly non-smooth functions (e.g., $\nu\ll d$, so that $\Delta$ being only roughly $O(T^{\frac{1}{3}})$).  

% To illustrate a more practical example, consider optimizing an one-dimensional (i.e., $d=1$) stock curve, which is a common financial challenge.  The erratic nature of stock prices often calls for the Mat\'ern kernel with $\nu=1/2$, which mimics the Brownian motion.  However, assuming an accumulated stock change of $\Delta=\sqrt{T}$ over time $T$, the upper bound from \cite{Hon23} yields a linear regret.  This limitation highlights the need for further improvement of the Mat\'ern upper bound.

% \medskip
{\bf \noindent Contributions. } We are unaware of any existing lower bounds in the non-stationary kernelized setting, leaving it unclear to what extent the upper bounds can be improved.  To address this, we establish lower bounds building on techniques from stationary settings \cite{Bul11,Sca17,Cai21}, focusing on the widely-adopted Mat\'ern kernel whose parameter $\nu$ can be varied to produce very rough functions (e.g., $\nu = \frac{1}{2}$) or very smooth ones (for large $\nu$).  By comparing our lower bounds to the upper bound in \eqref{eq:RT_Matern} (along with the trivial $O(T)$ upper bound), we find that despite gaps between the two, they are very close in a broad range of parameter regimes, including certain limits such as $\nu \to \infty$ but also finite values.  The gaps are especially narrow under the $\ell_{\infty}$ constraint (i.e., $\dagger = \infty$ in \eqref{eq:restriction}), and the larger gaps for $\dagger = k$ raise the interesting open question on whether there are any regimes in which $\dagger = k$ poses a strictly easier problem (i.e., lower scaling in the regret) than $\dagger = \infty$.

A summary of our regret bounds and the upper bounds of \cite{Hon23} are given in Table \ref{tbl:main}, and a more detailed comparison will be given in Section \ref{sec:cmp}.

\rev{
    {\bf Note on Concurrent Work.} A concurrent work \cite{iwazaki2024near} also derived our lower bound for $\ell_{\infty}$-norm variation, and moreover, they gave an algorithm having a matching upper bound to within logarithmic factors.  We still discuss the gaps to the upper bounds in \cite{Hon23} throughout this paper, which remain of interest even following the work of \cite{iwazaki2024near}.  We note that RKHS norm variation was not considered in \cite{iwazaki2024near}.
}

\begin{table*}
\begin{center}
\begin{tabular}{|c|c|c|c|}
	\hline 
	& $L$ switches & $\ell_{\infty}$ variation & \phantom{$\bigg(^{()}$} RKHS variation \tabularnewline
	\hline 
	\hline 
	Upper Bound (Existing) & $\widetilde{O}\big(\sqrt{L}T^{\frac{\nu+d}{2\nu+d}}\big)$ & $\widetilde{O}\big(\Delta^{\frac{1}{3}}T^{\frac{4\nu/3+d}{2\nu+d}}\big)$ & \phantom{$\bigg(^{()}$} $\widetilde{O}\big(\Delta^{\frac{1}{3}}T^{\frac{4\nu/3+d}{2\nu+d}}\big)$ \tabularnewline
	\hline 
	Lower Bound (Ours) & $\Omega\big(L^{\frac{\nu}{2\nu+d}}T^{\frac{\nu+d}{2\nu+d}}\big)$ & $\Omega\big(\Delta^{\frac{\nu}{3\nu+d}}T^{\frac{2\nu+d}{3\nu+d}}\big)$ & $\begin{cases}
		\Omega\big(\Delta^{\frac{\nu}{2\nu+d}}T^{\frac{\nu+d}{2\nu+d}}\big) & \nu\le d\\
		\Omega\big(\Delta^{\frac{1}{3}}T^{\frac{2}{3}}\big) & \nu\ge d
	\end{cases}$ \tabularnewline
	\hline 
\end{tabular}
\caption{Summary of our lower bounds and the upper bounds from \cite{Hon23}. \label{tbl:main}}
\end{center}
\end{table*}

% While the SE kernel result from \cite{Hon23} is nearly optimal (i.e., up to logarithmic factors), there remains a polynomial gap for the Mat\'ern kernel.  More specifically, we provide a $\Omegatil(\Delta^{\frac{\nu}{3\nu+d}} T^{\frac{2\nu+d}{3\nu+d}})$ lower bound in the case of the Mat\'ern kernel, indicating potential areas for improvement.

% Furthermore, we demonstrate an algorithm that can achieve a regret of at most $ \Otil(\Delta^{\frac{2\nu}{6\nu+3d}} T^{\frac{4\nu+3d}{6\nu+3d}})$ for the Mat\'ern kernel, which, to our knowledge, represents the first guaranteed sub-linear Mat\'ern regret in the time varying setting.  Our algorithm is inspired by the batched elimination algorithms in the stationary setting (e.g., see \cite{Cam21,Li22}), and utilizes two nested batches to detect environment changes in nearly real-time.  See Section \ref{sec:ub} for more details.
% Thus, using the RKHS norm as the budget could potentially simplify the problem compared to using the infinity norm.  For example, there are cases where the change in RKHS norm can be substantial even if the infinity norms change slightly, so that the algorithm dose not need to adapt a lot since the RKHS budget $\delk$ will be quickly exhausted.  These considerations are reflected in our theoretical results for $\dagger=k$.

\section{Related Work}\label{sec:relate}
% Across classic bandits, convex function optimization, and kernel function optimization, researchers have explored diverse approaches with theoretical guarantees to tackle challenges arising from non-stationary environments.

A variety of non-stationary bandit problems have been considered previously; we give a brief outline as follows without seeking to be exhaustive.

% \medskip
{\bf\noindent Finite-Arm Bandits.}  Problems of non-stationary bandits in finite-arm scenarios have been studied in detail.  For example, \cite{Gar11} formulated the problem as {\em switching bandits} or abruptly-changing environments, bounding the total number of changes by $L\ge 1$.  A subsequent line of research, including \cite{Bes14, Che19}, proposed the use of variation bounds similar to \eqref{eq:restriction} to allow both slowly-varying and abruptly-changing environments.  

Near-optimal algorithms and matching lower bounds have been developed for $K$-armed bandits \cite{Bes14, Aue19} in order of $\Theta(K^{\frac{1}{3}}\Delta^{\frac{1}{3}} T^{\frac{2}{3}})$. %and linear bandits \cite{Che19, Rus19} in order of $\Theta(d^{\frac{2}{3}}\Delta^{\frac{1}{3}} T^{\frac{2}{3}})$.  
Notably, the meta-algorithm introduced in \cite{Wei21}, called the MASTER reduction, offers a general multi-scale sampling framework that achieves near-optimal regret guarantees for finite bandit problems, among others.
% holding the potential for extension to function optimization.

% \medskip
{\bf\noindent Convex Function Optimization.} Time-varying bandit convex optimization is considered \cite{Bes15} with prior knowledge of the changing budget $\Delta$.  Their work leverages a restarting policy, and establishes that $\Theta(d^{\frac{1}{3}} \Delta^{\frac{1}{3}} T^{\frac{2}{3}})$ is information-theoretically optimal.  Subsequent works, such as \cite{Bab21,Bab22}, generalize this result to cases without knowledge of $\Delta$.  The MASTER reduction can also be combined with \cite{Wan23} to attain a similar guarantee.
% As hinted, the MASTER backbone is then combined by \cite{Wan23} to achieve another optimal regret with UCB-type algorithms.

% \medskip
{\bf\noindent Linear Bandits.} Non-stationary linear bandits are an important special case of non-stationary kernelized bandits, with functions of the form $f_t(\xv) = \btheta_t^T \xv$.  In the case that $\dagger = k$ in \eqref{eq:restriction}, the variation constraint simplifies to $\sum_{t=1}^{T-1} \|\btheta_{t+1}-\btheta_{t}\|_{2} \le \Delta$.  For this setting, the optimal scaling of cumulative regret was shown in \cite{cheung2022hedging} to be $\widetilde{\Theta}(d^{2/3} \Delta^{1/3} T^{2/3} )$ (where the lower bound restricts certain scalings in a manner that prevents a contradiction with a trivial $O(T)$ upper bound).  The upper bound in their work required knowledge of $\Delta$, and subsequently \cite{Hon23} obtained the same upper bound (to within logarithmic factors) without requiring such knowledge.

% \medskip
{\bf\noindent Kernelized Bandits.} Before discussing the kernelized setting, we re-iterate that it is customary to take the dimension $d$ as being constant in such settings, unlike convex and linear settings where the $d$ dependence is maintained in the results.  This is because kernelized bandits with standard choices of the kernel suffer a curse of dimensionality in high dimensions, and overcoming this curse (e.g., by using a kernel with low-dimensional structure) is typically treated as a separate topic.

Initial studies of non-stationary kernel bandits (focusing on $\dagger = k$) considered techniques such as sliding window \cite{Zho21} and penalty discounting \cite{Den22}.  These works establish an upper bound of $\Otil(\gamma_T^{\frac{7}{8}}\Delta^{\frac{1}{4}}T^{\frac{3}{4}})$ with prior knowledge of $\Delta$, which may not be sublinear in $T$ even when $\Delta$ is a constant.  As mentioned in the introduction, \cite{Hon23} gives a significantly improved upper bound of $\Otil\big( \gamma_T^{\frac{1}{3}} \Delta^{\frac{1}{3}} T^{\frac{2}{3}}\big)$ even without knowledge of $\Delta$, and even with $\dagger = \infty$.

With unknown $\Delta$, the versatile MASTER reduction \cite{Wei21} can also readily be extended to the kernelized setting when coupled with the foundational GP-UCB \cite{Sri09}.  This combination, as demonstrated in \cite[App.~E]{Hon23}, yields a regret bound of $\Otil(\gamma_T \Delta^{\frac{1}{3}} T^{\frac{2}{3}})$.  We show in Appendix \ref{app:master} that even with a ``best conceivable'' variant of GP-UCB, this combination would only see a slight improvement to $\Otil(\gamma_T^{\frac{2}{3}} \Delta^{\frac{1}{3}} T^{\frac{2}{3}})$.  %,  though  this maintains consistency with linear bandit results in replacing dimension $d$ with the ``effective dimension'' $\gamma_T$.  
This reveals a possible fundamental limitation of the MASTER approach (at least with the tools currently available) that prevents it from matching the $\Otil(\gamma_T^{\frac{1}{3}} \Delta^{\frac{1}{3}} T^{\frac{2}{3}})$ scaling in \cite{Hon23}.

Another related work is \cite{Bog16}, in which a distinct \emph{Bayesian} setting is considered with a slowly-varying function.  This is generally less related to our work, as we consider \emph{frequentist} (RKHS) modeling assumptions, and the variation budget in \eqref{eq:restriction} allows for both gradual changes and abrupt changes.

% \medskip
{\bf\noindent Lower Bounds for Stationary Kernelized Bandits.} In the stationary setting, under the Mat\'ern kernel with parameter $\nu$, a regret lower bound of $\Omega\big( T^{\frac{\nu+d}{2\nu+d}} \big)$ was shown in \cite{Sca17} (see also \cite{Cai21}).  Up to logarithmic factors, this matches the $\Otil(\sqrt{T \gamma_T})$ upper bound first achieved via SupKernelUCB \cite{Val13} and later via other algorithms \cite{Li22,salgia2020,jamieson21}.  Note that $\frac{\nu+d}{2\nu+d} \to \frac{1}{2}$ as $\nu \to \infty$, indicating a dependence approaching $\sqrt{T}$.  Analogous kinds of lower bounds have since been established for other kernel-based setting, e.g., with robustness considerations \cite{Cai21,Bog18} or heavy-tailed noise \cite{Cho19}.

 % and kernel function optimization, we observe that for less restricted kernel functions, such as Mat\'ern-$\nu$ functions, the optimal scaling depends on the smoothness $\nu$ and dimension $d$, manifesting as $T^{\frac{2\nu+d}{3\nu+d}}$.  

\section{Problem Setup} \label{sec:setup}

We consider the problem of optimizing a dynamic RKHS black-box function $f_t$ over a compact domain $D=[0,1]^d$, whose RKHS norm is bounded as $\|f_t\|_k\le B$ for all $t = 1,\dotsc,T$.  At each time $t$, the player chooses an action $\xv_t\in D$, and receives a noisy observation 
\begin{equation}
	y_t = f_t(\xv_t)+z_t,
\end{equation}
where the noise terms $z_t\sim \Nc(0,\sigma^2)$ are i.i.d.~across time.  The performance is measured using the {\em dynamic} cumulative regret:
\begin{equation}
    R_T = \sum_{t=1}^{T} \big( f_t(\xv^*_t) - f_t(\xv_t) \big),
\end{equation}
where $\xv_t^* = \argmax_{\xv\in D} f_t(\xv)$ maximizes $f_t(\xv)$.

As stated in the introduction, the amount of function variation is assumed to be bounded by a value $\Delta$ according to either the $\ell_{\infty}$-norm or the RKHS norm:
\begin{itemize}
	\item For the $\ell_{\infty}$-norm ($\dagger = \infty$):
	\begin{equation}
		\sum_{t=1}^{T-1} \|f_{t+1}-f_{t}\|_{\infty} \le \Delta. \label{eq:var1}
	\end{equation}
	\item For the RKHS norm ($\dagger = k$):
	\begin{equation}
		\sum_{t=1}^{T-1} \|f_{t+1}-f_{t}\|_{k} \le \Delta. \label{eq:var2}
	\end{equation}
\end{itemize}
We will also briefly consider the case of a limited number of switches:
\begin{equation}
	\sum_{t=1}^T \openone \{f_{t+1}\neq f_t \} \le L-1, \label{eq:var3}
\end{equation}
but we will give this less attention due to being much easier to handle.

Previous studies of lower bounds in kernelized bandits have focused on the Mat\'ern and squared exponential (SE) kernels.  While our techniques apply to both, we focus on the Mat\'ern since it is more versatile via its smoothness parameter, with the limit $\nu \to \infty$ essentially capturing the SE kernel behavior up to minor differences in logarithmic terms.  \rev{We provide some further discussion on the extension to the SE kernel in Appendix \ref{app:SE}.}

Formally, the Mat\'ern kernel is described as follows:
\begin{align}
	k(x,x') &= \dfrac{2^{1-\nu}}{\Gamma(\nu)} \bigg(\dfrac{\sqrt{2\nu}\|x - x'\|}{\ell}\bigg)^{\nu}  J_{\nu}\bigg(\dfrac{\sqrt{2 \nu}\|x - x'\|}{\ell} \bigg), \label{eq:kMat}
\end{align}
where $\ell>0$ denotes the length-scale, $\nu > 0$ is an additional parameter that dictates the smoothness,  and $J_{\nu}$ denotes the modified Bessel function.

%We will focus on the regime $\Delta = o(T)$ (or $L = o(T)$), since if $\Delta = \Theta(T)$ it is straightforward to show that the cumulative regret must be linear in $T$, i.e., the standard goal of sub-linear regret is impossible.  
We will treat the parameters $(d,B,\sigma,\nu,\ell)$ as constants, and focus our attention on how the regret depends on $T$ and $\Delta$.  This is partially justified as follows: (i) It is natural to fix the function class in asymptotic analyses, and (ii) To our knowledge, optimal dependencies on these parameters are not even known in the stationary setting, so they would deserve a more detailed treatment there before addressing them in the time-varying setting. 

\section{Main Results}

In this section, we present our three main results in succession in Section \ref{sec:statements}, and then provide detailed comparisons to the respective upper bounds in Section \ref{sec:cmp}.  The proofs are outlined in Section \ref{sec:outline}, with the full details given in the supplementary material.  

\subsection{Theorem Statements} \label{sec:statements}
 
Our first result focuses on the case of a bounded number of changes as per \eqref{eq:L_changes}, which turns out to be easy to derive a lower bound for.

\begin{thm} \label{thm:L}
	{\bf (Limited Number of Changes)} Consider the setup in which the function changes at most $L-1$ times according to \eqref{eq:L_changes}, and suppose that the RKHS norm $B$, noise level $\sigma$, dimension $d$, and Mat\'ern parameters $(\nu,\ell)$ are all constant.  Then, with time horizon $T$ and $L \in \{1,\dotsc,T\}$, any algorithm must incur average cumulative regret $R_T$ satisfying
	\begin{equation}
		\EE[R_T] \ge \Omega\Big( L^{\frac{\nu}{2\nu+d}} T^{\frac{\nu+d}{2\nu + d}} \Big). \label{eq:bound_L}
	\end{equation}
	Moreover, $\EE[R_T] \ge \Omega( \sqrt{LT} )$ for all $(\nu,d)$.
\end{thm}

% {\color{blue}Compared to the upper bound, we have $L^{\frac{\nu}{2\nu + d}}$ dependence instead of $\sqrt{L}$, so the two match when $L = O(1)$ or when $\nu$ grows large.  Otherwise there may be a gap, e.g., when $\nu = d$ it's $L^{1/3}$ vs.~$L^{1/2}$.}

Next, we present our results for the norm variation condition \eqref{eq:restriction}, considering $\dagger = \infty$ and $\dagger = k$ separately.

\begin{thm} \label{thm:Inf}
	{\bf (Limited $\ell_{\infty}$-Norm Variation)} Consider the setup with $\ell_{\infty}$-norm variation at most $\Delta$ according to \eqref{eq:restriction} with $\dagger = \infty$, and suppose that the norm bound $B$, noise level $\sigma$, dimension $d$, and Mat\'ern parameters $(\nu,\ell)$ are all constant.  Then, with time horizon $T$, any algorithm must incur average cumulative regret $R_T$ satisfying
	\begin{equation}
		\EE[R_T] \ge \Omega\Big( \Delta^{\frac{\nu}{3\nu+d}} T^{\frac{2\nu+d}{3\nu + d}} \Big). \label{eq:bound_Inf}
	\end{equation}
	Moreover, $\EE[R_T] \ge \Omega( \Delta^{1/3} T^{2/3} )$ for all $(\nu,d)$.
\end{thm}

\begin{thm} \label{thm:rkhs}
	{\bf (Limited RKHS Norm Variation)} Consider the setup with RKHS norm variation at most $\Delta$ according to \eqref{eq:restriction} with $\dagger = k$, and suppose that the norm bound $B$, noise level $\sigma$, dimension $d$, and Mat\'ern parameters $(\nu,\ell)$ are all constant.  Then, with time horizon $T$, any algorithm must incur average cumulative regret $R_T$ satisfying the following:
	\begin{itemize}
		\item[(i)] If $\nu \le d$,\footnote{The case that $\nu = d$ is included in both cases since in that scenario the bounds in \eqref{eq:bound_k1} and \eqref{eq:bound_k2} become identical.} then
		\begin{equation}
			\EE[R_T] \ge \Omega\Big( \Delta^{\frac{\nu}{2\nu+d}} T^{\frac{\nu+d}{2\nu + d}} \Big). \label{eq:bound_k1}
		\end{equation}
		\item[(ii)] If $\nu \ge d$, then 
		\begin{equation}
			\EE[R_T] \ge \Omega\Big( \Delta^{1/3}T^{2/3} \Big), \label{eq:bound_k2}
		\end{equation}
	\end{itemize}
	Moreover, $\EE[R_T] \ge \Omega( \Delta^{1/3} T^{2/3} )$ for all $(\nu,d)$.
\end{thm}

Before giving a detailed comparison to the upper bounds from \cite{Hon23}, we highlight some desirable properties of our lower bounds:
\begin{itemize}
	\item The $T$-dependence is always at least as high as that of the standard setting, which is $T^{\frac{\nu+d}{2\nu+d}}$ \cite{Sca17}, and in \eqref{eq:bound_Inf} and \eqref{eq:bound_k2} a strictly higher $T$-dependence is observed (analogous to $\sqrt{T}$ vs.~$T^{2/3}$ for linear bandits).  We note that the standard lower bound trivially still applies in our setup, so \eqref{eq:bound_Inf}--\eqref{eq:bound_k2} all remain valid when $\Omega(T^{\frac{\nu+d}{2\nu+d}})$ is added to the right-hand side (such a term may become dominant when $\Delta = o(1)$).
	\item As $\nu \to \infty$ we approach the dependencies on $\Delta$ and $T$ observed in the linear bandit setting (namely, $\Delta^{1/3} T^{2/3}$), which is consistent with the fact that highly smooth RKHS functions (e.g., the SE kernel) typically come with similar bounds as linear bandits \cite{Sri09,Hon23}.
	\item If $L = T^\beta$ or $\Delta = T^\beta$ with $\beta \in (0,1)$, then we get $\EE[R_T] \ge \Omega(T^{1-\epsilon})$ for arbitrarily small $\epsilon$ as $\beta$ approaches one (for any fixed $\nu$ and $d$).  This matches the expectation that when the amount of variation becomes close to linear, the regret should also become close to linear.  Similarly, if $\Delta = \Theta(T)$ (or $L = \Theta(T)$) then $R_T = \Omega(T)$.
\end{itemize}

\subsection{Comparisons to Upper Bounds} \label{sec:cmp}

Recall that our lower bounds are compared to the existing upper bounds in Table \ref{tbl:main}.  In this subsection, we provide a more detailed comparison.

For the case of a limited number of changes, the comparison between the upper bound \eqref{eq:RT_Matern_L} and our lower bound \eqref{eq:bound_L} is relatively straightforward.  The two have the same $T$-dependence $T^{\frac{\nu+d}{2\nu+d}}$ (also matching the standard setting \cite{Sca17}), whereas the $L$ dependence differs in being $\sqrt{L}$ vs.~$L^{\frac{\nu}{2\nu+d}}$.  Thus, the two are very close when $\nu$ is large, but may be less so when $\nu$ is small (e.g., $\sqrt{L}$ vs.~$L^{1/3}$ when $\nu=1$).  While the gap may seem large when $\nu \ll d$, this is in fact another regime in which our lower bound becomes tight, as the $T$ dependence itself becomes close to linear, thus nearly matching a trivial upper bound of $O(T)$.

For the case of bounded variation according to $\Delta(\infty)$ or $\Delta(k)$, the comparison becomes less straightforward due to both the $\Delta$ and $T$ dependencies differing (except in \eqref{eq:bound_k2} where we match the $\Delta^{1/3}$ dependence and differ by a multiplicative $T^{\frac{d/3}{2\nu+d}}$ term in the $T$-dependence).  Thus, instead of an analytical comparison, we find it more insightful to set $\Delta = \Theta(T^{\beta})$ for various choices of $\beta$, and compare the resulting values of $\alpha$ such that the regret has dependence $T^{\alpha}$.  In cases where \eqref{eq:RT_Matern} suggests setting $\alpha > 1$, we instead set $\alpha = 1$ in view of the trivial $O(T)$ upper bound.

\begin{figure*}[!t]
	{\centering \includegraphics[width=\mywidth]{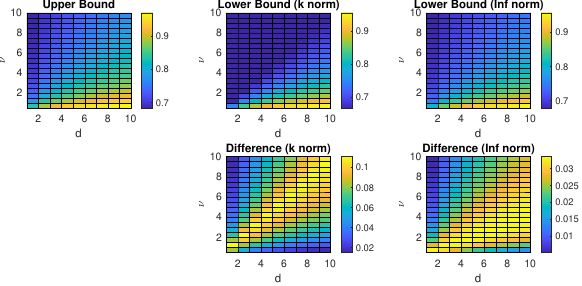} \par}
	\caption{Comparison of upper and lower bounds when $\Delta = \Theta(1)$.  (Top) Values of $\alpha$ such that the regret has dependence $T^{\alpha}$.  (Bottom) Differences of the upper bound's $\alpha$ value and the lower bound's value.  \emph{For $\ell_{\infty}$-norm, the highest difference is below 0.035}. \label{fig:a0}}
\end{figure*}

\begin{figure*}[!t]
	{\centering \includegraphics[width=\mywidth]{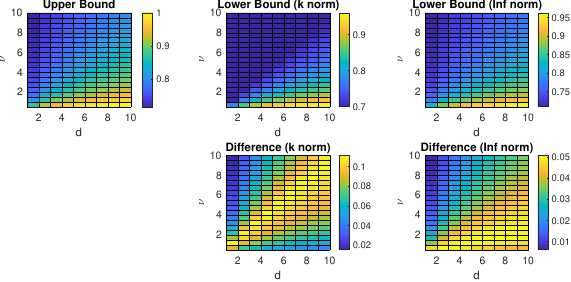} \par}
	\caption{Comparison of upper and lower bounds when $\Delta = \Theta(T^{0.1})$.  (Top) Value $\alpha$ such that the regret has dependence $T^{\alpha}$.  (Bottom) Differences of the upper bound's $\alpha$ value and the lower bound's value.  \emph{For $\ell_{\infty}$-norm, the highest difference is around 0.05}. \label{fig:a01}}
\end{figure*}

\begin{figure*}[!t]
	{\centering \includegraphics[width=\mywidth]{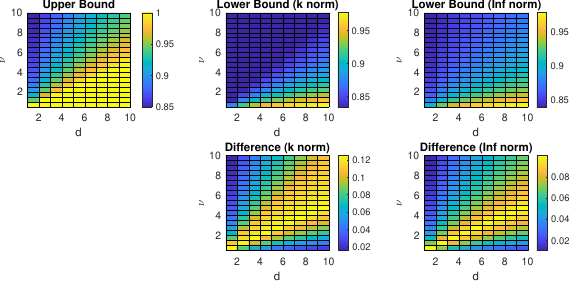} \par}
	\caption{Comparison of upper and lower bounds when $\Delta = \Theta(\sqrt{T})$.  (Top) Value $\alpha$ such that the regret has dependence $T^{\alpha}$.  (Bottom) Differences of the upper bound's $\alpha$ value and the lower bound's value.  \emph{For $\ell_{\infty}$-norm, the highest difference is below 0.1}. \label{fig:a05}}
\end{figure*}

\begin{figure*}[!t]
	{\centering \includegraphics[width=\mywidth]{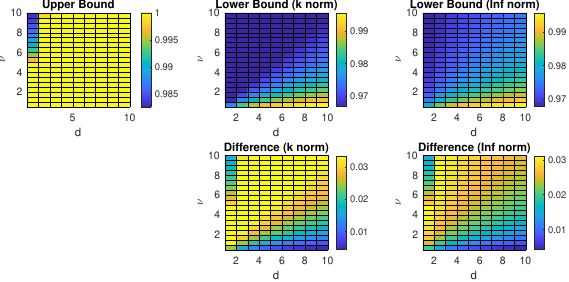} \par}
	\caption{Comparison of upper and lower bounds when $\Delta = \Theta(T^{0.9})$.  (Top) Value $\alpha$ such that the regret has dependence $T^{\alpha}$.  (Bottom) Differences of the upper bound's $\alpha$ value and the lower bound's value.  \emph{For $\ell_{\infty}$-norm, the highest difference is around 0.03}. \label{fig:a09}}
\end{figure*}

\begin{figure*}[!t]
    {\centering \includegraphics[width=\mywidth]{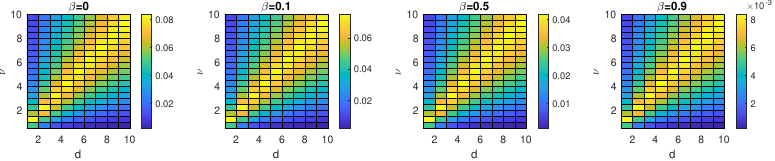} \par}
    \caption{Comparison of lower bounds for $\ell_{\infty}$-norm variation and RKHS norm variation for various $\beta$ values such that $\Delta = \Theta(T^{\beta})$.  Specifically, the difference is between the two values of $\alpha$ for which the regret has dependence $T^{\alpha}$. \label{fig:k_inf_cmp}}
\end{figure*}

The results are shown in Figures \ref{fig:a0}--\ref{fig:a09} for $\beta \in \{0,0.1,0.5,0.9\}$ respectively.  We first discuss the case of $\ell_{\infty}$-norm variation (i.e., $\dagger = \infty$ in \eqref{eq:restriction}):
\begin{itemize}
	\item For small $\beta$ (Figures \ref{fig:a0} and \ref{fig:a01}), the regret bounds are quite accurate, with the $\alpha$ values in the upper and lower bounds differing by at most $0.035$ for the $(d,\nu)$ pairs shown when $\Delta=\Theta(1)$, and around $0.05$ or less when $\Delta = \Theta(T^{0.1})$.  
	\item For large $\beta$ (Figure \ref{fig:a09}), the lower bound is again accurate due to nearly matching the trivial $O(T)$ (i.e, $\alpha=1$) upper bound.  In particular, with $\Delta = \Theta(T^{0.9})$ the lower bound's $\alpha$ value is consistently around $0.97$ or higher.
	\item The case of moderate $\beta$ (Figure \ref{fig:a05}) is where we see the largest gap between the upper and lower bounds.  However, for $\beta = 0.5$ we still see a gap of at most $0.1$ in the $\alpha$ values for the $(d,\nu)$ pairs shown, which is not too large of a gap given that the values being compared are all above $0.8$.
	\item For small and moderate $\beta$, the largest gaps tend to be observed when $\nu$ is slightly smaller than $d$.
\end{itemize}
For RKHS norm variation (i.e., $\dagger = k$ in \eqref{eq:restriction}), the gaps are noticeably larger (namely, up to around $0.11$) for small $\beta$, and tend to be largest when $d \approx \nu$.  The behavior then becomes more comparable to $\dagger = \infty$ for moderate $\beta$, and the two become very similar for high $\beta$.  \rev{These observations can also be seen from Figure \ref{fig:k_inf_cmp}, where we plot the difference between the $\alpha$ values for $\dagger = \infty$ and $\dagger = k$.} Overall, these findings raise the interesting question of \emph{whether improved upper bounds for the case $\dagger = k$ are possible, particularly when $\Delta=O(T^{\beta})$ for relatively small $\beta$ values.}

\rev{We proceed to give some intuition as to why the gaps tend to be larger when $\nu$ and $d$ are comparable (and often roughly equal):
\begin{itemize}
    \item  As $\frac{d}{\nu} \to \infty$, the lower bounds become nearly linear in $T$, thus nearly matching the trivial $O(T)$ upper bound.
    \item As $\frac{d}{\nu} \to 0$, the upper and lower bounds both have $T^{2/3}$ dependence, thus again implying a small gap.
\end{itemize}
Thus, the largest gap should tend to occur when $\frac{d}{\nu}$ is neither too small nor too large.   Moreover, since all of the bounds only depend on $(\nu,d)$ via their ratio $\frac{\nu}{d}$, it is reasonable to expect that there exists a ``worst-case'' ratio where the gap is maximized (though this ratio can vary with $\beta$).
}

Finally, we note that the parts of our theorem statements regarding arbitrary $(\nu,d)$ pairs (in particular, arbitrarily large $\nu$) have scaling laws matching those of linear bandits in constant dimension \cite{Hon23}, from which we can conclude (as one would expect) that the kernelized setting with the Mat\'ern kernel is at least as hard as the linear bandit setting.  As $\frac{\nu}{d}$ increases, the gap in difficulty between the two settings becomes smaller. %; specifically, the multiplicative gap is upper bounded by $T^{\epsilon(\nu/d)}$ for some $\epsilon(\nu)$ satisfying $\lim_{\nu \to \infty} \epsilon(\nu) = 0$.

\subsection{Outline of Proofs} \label{sec:outline}

The proofs of Theorems \ref{thm:L}, \ref{thm:Inf}, and \ref{thm:rkhs} (given in the supplementary material) have increasing complexity, but all share common ideas.  In all cases, we build on the lower bound for the stationary setting \cite{Sca17}, in which $M$ functions are formed each having a ``bump'' in a distinct location, and successful location requires locating the bump.  With a bump height of $2\epsilon$, we require $\Omega\big( \frac{M}{\epsilon^2} \big)$ queries to locate it, and accordingly, if $T \le O\big( \frac{M}{\epsilon^2} \big)$ with a small enough implied constant then cumulative regret $\Omega(T \epsilon)$ is incurred.  By carefully choosing $M$ and $\epsilon$ according to the Mat\'ern smoothness assumption (which prevents the bump from becoming too narrow), this approach gives a cumulative regret bound of $\Omega\big( T^{\frac{\nu+d}{2\nu+d}} \big)$ for the stationary setting.

{\bf Outline for Theorem \ref{thm:L} ($L-1$ Switches).} This result follows easily from the stationary lower bound just mentioned: We simply divide the time horizon into $L$ regions of length $\frac{T}{L}$, have the function change arbitrarily from one region to the next, and apply the standard lower bound in each region.

{\bf Outline for Theorem \ref{thm:Inf} ($\dagger = \infty$).} We similarly divide the time horizon into $c$ regions of length $\frac{T}{c}$, with the function changing from one region to the next, but now we need to choose $c$ to ensure that the $\ell_{\infty}$-norm variation constraint \eqref{eq:restriction} is satisfied.  Since the bump functions that we use have $\ell_{\infty}$-norm $2\epsilon$, it suffices to choose $c = \frac{\Delta}{4\epsilon}$.  Combining this equation with an expression for $\epsilon$ in terms of $\frac{T}{c}$ (coming from the steps of the standard lower bound proof), we can deduce an expression for $\epsilon$ in terms of $(T,\Delta)$ alone.  The cumulative regret incurred is then on the order of $c \cdot \frac{T}{c} \cdot \epsilon$ (i.e., $\frac{T}{c} \cdot \epsilon$ in each of the $c$ regions), and substituting the expression for $\epsilon$ gives the desired result.

{\bf Outline for Theorem \ref{thm:rkhs} ($\dagger = k$).} As before, we divide the time horizon into $c$ regions of length $\frac{T}{c}$. In this case, further care is we need to balance between two sources of $\|f_t\|_k$ being bounded, one being the direct bound $\|f_t\|_k \le B$ and the other being $\|f_t\|_k \le \frac{\Delta}{2c}$ to satisfy the constraint \eqref{eq:restriction}.  Unlike in the first two theorems, we find that setting $\|f_t\|_k = B$ does not always give the best lower bound.

If we do set $\|f_t\|_k = B$, we find that the remaining analysis becomes similar to that of Theorem \ref{thm:Inf}, but with a lower bound of $\Omega\big(\Delta^{\frac{\nu}{2\nu + d}} T^{\frac{\nu+d}{2\nu + d}}\big)$ instead of $\Omega\big( \Delta^{\frac{\nu}{3\nu+d}} T^{\frac{2\nu+d}{3\nu + d}} \big)$ due to the different norms being used.  Since $\Delta \le T$, the former threshold is lower, and this stems from the fact that each function change uses up a variation budget of $O(B) = O(1)$ under the RKHS norm, but only $O(\epsilon) = o(1)$ under the $\ell_{\infty}$-norm.

We can, however, do better by setting $\|f_t\|_k = \frac{\Delta}{2c}$ and allowing this to be much smaller than $B$.  Interestingly, upon doing so, we get a regret lower bound depending on $c$ such that:
\begin{itemize}
	\item If $d > \nu$ then the bound gets better as $c$ decreases;
	\item If $d < \nu$ then the bound gets better as $c$ increases.
\end{itemize}
While the most obvious constraint on $c$ is that $c \in \{1,\dotsc,T\}$, we are in fact further constrained by requiring the functions used to have small $\ell_{\infty}$-norm for the lower bounding methods from \cite{Sca17} to be applicable, and by requiring the choice $\|f_t\|_k = \frac{\Delta}{2c}$ to remain consistent with $\|f_t\|_k \le B$.  Upon choosing the smallest/largest $c$ values permitted by these conditions, we find that (i) when $d > \nu$, it is best to stick to the approach taken in the case that $\|f_t\|_k = B$ (see above), which yields \eqref{eq:bound_k1}, and (ii) when $d < \nu$, taking the largest permissible choice of $c$ gives the improved bound in \eqref{eq:bound_k2}.

\section{Open Problems} \label{sec:open}

While our results generally show a good match between our lower bounds and the upper bounds of \cite{Hon23} (and in same cases, the trivial upper bound $O(T)$), it is worth noting a number of interesting open problems.

\rev{In an earlier version of this paper, we noted the open problems of closing the remaining gaps further, and deriving an upper bound that is sublinear in $T$ for all $(\nu,d)$ whenever $\Delta = o(T)$.  As we noted in the introduction, these problems were resolved in a concurrent work \cite{iwazaki2024near} showing that our lower bound for $\ell_{\infty}$-norm variation is tight to within logarithmic factors.

To our knowledge, the following two further problems still remain open:}

\begin{itemize}
	% \item[(i)] \emph{To what extent can the gaps be closed further?}  Answering this might involve sharpening the analysis of the algorithms in \cite{Hon23} or coming up with new algorithms, or similarly sharpening the analysis of the lower bounds of the function classes we satisfy or coming up with new hard function classes.  We expect that both our analysis and that of \cite{Hon23} do not leave much room for being sharpened, but rather more distinct new ideas will be needed.
	% \item[(ii)] {\em Is it possible to attain an upper bound that is sublinear in $T$ for all $(\nu,d)$ whenever $\Delta = o(T)$?} (Or $L=o(T)$ in the case of finitely many switches.)  The bound in \eqref{eq:RT_Matern} does not satisfy this requirement, as it is only sublinear when $\Delta^{1/3} \ll T^{\frac{2\nu/3}{2\nu+d}}$, or equivalently $\Delta \ll T^{ \frac{2\nu}{2\nu+d} }$ (e.g., $\Delta \ll T^{2/3}$ when $\nu = d$).
	\item[(i)] {\em Are there any regimes in which the best possible regret with $\dagger = k$ is strictly smaller than that of $\dagger = \infty$?}  \rev{The above-mentioned work \cite{iwazaki2024near} focuses on the case that $\dagger = \infty$, whose upper bounds imply the same upper bounds for $\dagger = k$.  However, our lower bounds suggest that there could be room for further improvement when $\dagger = k$.}
	\item[(ii)] {\em For the case that $\dagger = k$, is the $\Omega\big( \Delta^{1/3} T^{2/3} \big)$ lower bound tight (to within logarithmic factors) for certain finite $(\nu,d)$ pairs with $\nu \ge d$?}  The current upper bound only matches this in the limit as $\nu \to \infty$.  Alternatively, it may be that an improved lower bound can be derived showing that this goal is impossible for any finite values.
\end{itemize}

\section{Conclusion}

We have established, to our knowledge, the first lower bounds on regret for time-varying kernelized bandits under a constraint on the total function variation (or number of switches).  These bounds come close to the upper bounds \cite{Hon23} in broad scaling regimes of interest, though also give rise to a number of directions for further work, as detailed in Section \ref{sec:open}.

\rev{{\bf Acknolwedgment.} This research is supported by the Singapore National Research Foundation under its Global AI Visiting Professorship program.}

% \newpage
\bibliography{ref}
\bibliographystyle{apalike}

%%%%%%%%%%%%%%%%%%%%%%%%%%%%%%%%%%%%%%%%%%%%%%%%%%%%%%%%%%%%
\section*{Checklist}

% %%% BEGIN INSTRUCTIONS %%%
%The checklist follows the references. For each question, choose your answer from the three possible options: Yes, No, Not Applicable.  You are encouraged to include a justification to your answer, either by referencing the appropriate section of your paper or providing a brief inline description (1-2 sentences). 
%Please do not modify the questions.  Note that the Checklist section does not count towards the page limit. Not including the checklist in the first submission won't result in desk rejection, although in such case we will ask you to upload it during the author response period and include it in camera ready (if accepted).

%\textbf{In your paper, please delete this instructions block and only keep the Checklist section heading above along with the questions/answers below.}
% %%% END INSTRUCTIONS %%%

\begin{enumerate}

	\item For all models and algorithms presented, check if you include:
	\begin{enumerate}
		\item A clear description of the mathematical setting, assumptions, algorithm, and/or model. [Yes]
		\item An analysis of the properties and complexity (time, space, sample size) of any algorithm. [Yes]
		\item (Optional) Anonymized source code, with specification of all dependencies, including external libraries. [Not Applicable]
	\end{enumerate}

	\item For any theoretical claim, check if you include:
	\begin{enumerate}
		\item Statements of the full set of assumptions of all theoretical results. [Yes]
		\item Complete proofs of all theoretical results. [Yes]
		\item Clear explanations of any assumptions. [Yes]
	\end{enumerate}

	\item For all figures and tables that present empirical results, check if you include:
	\begin{enumerate}
		\item The code, data, and instructions needed to reproduce the main experimental results (either in the supplemental material or as a URL). [Not Applicable]
		\item All the training details (e.g., data splits, hyperparameters, how they were chosen). [Not Applicable]
		\item A clear definition of the specific measure or statistics and error bars (e.g., with respect to the random seed after running experiments multiple times). [Not Applicable]
		\item A description of the computing infrastructure used. (e.g., type of GPUs, internal cluster, or cloud provider). [Not Applicable]
	\end{enumerate}
	
	\item If you are using existing assets (e.g., code, data, models) or curating/releasing new assets, check if you include:
	\begin{enumerate}
		\item Citations of the creator If your work uses existing assets. [Not Applicable]
		\item The license information of the assets, if applicable. [Not Applicable]
		\item New assets either in the supplemental material or as a URL, if applicable. [Not Applicable]
		\item Information about consent from data providers/curators. [Not Applicable]
		\item Discussion of sensible content if applicable, e.g., personally identifiable information or offensive content. [Not Applicable]
	\end{enumerate}
	
	\item If you used crowdsourcing or conducted research with human subjects, check if you include:
	\begin{enumerate}
		\item The full text of instructions given to participants and screenshots. [Not Applicable]
		\item Descriptions of potential participant risks, with links to Institutional Review Board (IRB) approvals if applicable. [Not Applicable]
		\item The estimated hourly wage paid to participants and the total amount spent on participant compensation. [Not Applicable]
	\end{enumerate}
	
\end{enumerate}

%%%%%%%%%%%%%%%%%%%%%%%%%%%%%%%%%%%%%%%%%%%%%%%%%%%%%%%%%%%%%%%%%%%%%%%%%%%%%%%
%%%%%%%%%%%%%%%%%%%%%%%%%%%%%%%%%%%%%%%%%%%%%%%%%%%%%%%%%%%%%%%%%%%%%%%%%%%%%%%
% APPENDIX
%%%%%%%%%%%%%%%%%%%%%%%%%%%%%%%%%%%%%%%%%%%%%%%%%%%%%%%%%%%%%%%%%%%%%%%%%%%%%%%
%%%%%%%%%%%%%%%%%%%%%%%%%%%%%%%%%%%%%%%%%%%%%%%%%%%%%%%%%%%%%%%%%%%%%%%%%%%%%%%
\newpage
\appendix
\onecolumn

{\Huge \bf \centering Appendix \par}

\section{Note on Notation}

Throughout the appendix, we use $\asymp$, $\lesssim$, and $\gtrsim$ to represent $\Theta(\cdot)$, $O(\cdot)$, and $\Omega(\cdot)$ respectively.  Similarly, we write $A \ll B$ (or $B \gg A$) if $A = o(B)$.  Here the asymptotics are with respect to $T \to \infty$, where the variation parameter $\Delta$ (or $L$) may also scale with $T$.  In contrast, the parameters $(B,\sigma^2,d,\nu,\ell)$ are assumed to be constant, meaning we may omit their dependencies via statements such as $\sqrt{dT} \asymp \sqrt{T}$.

\section{Tools for Proving the Lower Bounds} \label{sec:tools}

\subsection{Stationary Setting}

We first introduce some tools from the stationary setting \cite{Sca17,Cai21} that will also be useful for our purposes.  The stationary setting corresponds to having $f_t = f, \forall t$ for a common function $f$ satisfying $\|f\|_k \le B$.  Throughout this section, we keep the dependence on $B$ explicit, since when we come to the non-stationary setting we will sometimes need to substitute different values that may depend on $T$.  (We still treat $\sigma^2$, $\nu$, $\ell$, and $d$ as constants.)

\begin{figure}
	\begin{centering}
		\includegraphics[width=0.7\columnwidth]{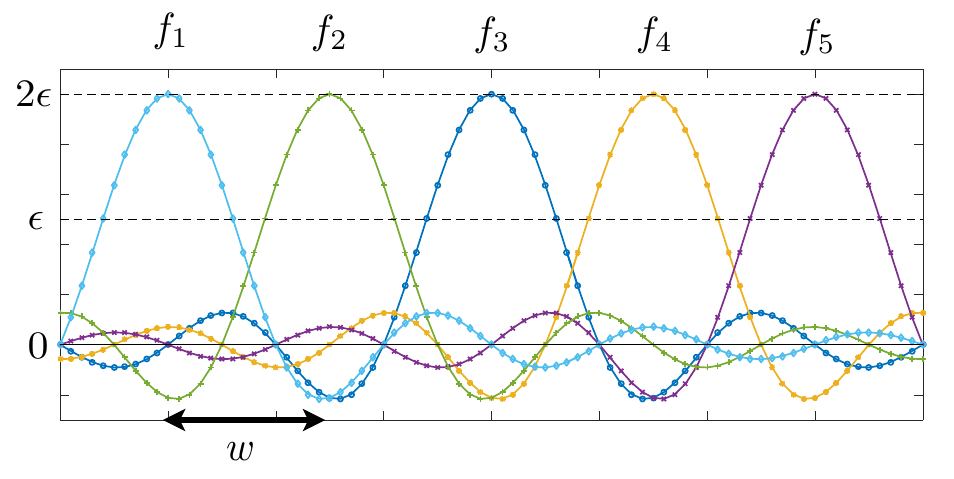}
		\par
	\end{centering}
	
	\caption{Illustration of five different ``bump'' functions, each having their peak in a different location.  Note that in this figure $f_1,\dotsc,f_5$ denotes five different stationary functions, as opposed to a non-stationary function at 5 different time indices. \label{fig:func_class}}
\end{figure}

We use the class of ``bump'' functions from \cite{Sca17}, illustrated in Figure \ref{fig:func_class}.  The idea is to consider $M$ functions, each equaling a fixed function $g(\xv)$ on $\RR^d$ shifted to have its peak in one of $M$ locations, and then cropped to the domain $[0,1]^d$.  The parameter $\epsilon > 0$ dictates the height bump, in particular ensuring that
\begin{equation}
	|g(\xv)| \le 2\epsilon ~~ \forall \xv, \quad \text{with equality at }\xv=\bzero. \label{eq:g_inf}
\end{equation}
The locations of the peaks are specified by a regular grid with width $w$, and the optimization problem becomes akin to finding a ``needle in a haystack'' among $M$ functions, where
\begin{equation}\label{eq:mw}
	M=\Big\lfloor \Big(\frac{1}{w} \Big)^d \Big\rfloor.
\end{equation}
It is established in \cite{Sca17} that these conditions, along with $\|g\|_k\le B$ (and thus $\|f\|_k \le B$) can be satisfied with a value of $M$ scaling as follows:
\begin{itemize} 
	\item For the squared exponential (SE) kernel,
	\begin{equation}\label{eq:m_se_infty}
		M = \Theta \Big(\Big(\log \frac{B}{\epsilon}\Big)^{\frac{d}{2}} \Big).
	\end{equation}
	\item For the Mat\'ern kernel,
	\begin{equation}\label{eq:m_mat_infty}
		M = \Theta \Big(\Big(\frac{B}{\epsilon}\Big)^{\frac{d}{\nu}} \Big).
	\end{equation}
\end{itemize}
In this paper, we focus on the Mat\'ern kernel due to its higher versatility.  For such functions, we could alternatively use the \emph{bounded support} bump from \cite{Bul11,Cai21} instead, but such choices would be infeasible for the SE kernel due to having infinite RKHS norm.

Note that the precise definition of $g$ will not be needed for our purposes; rather, all that we need is summarized in this subsection, particularly the lemmas below.  We henceforth let $\Fc_M$ denote the size-$M$ subset of hard functions.

The following lemmas state two results from \cite{Sca17} that we will use.  The second is their main result for the Mat\'ern kernel, and the first is an intermediate result that will also be useful.  We use a generic symbol $\tau$ for the time horizon, since in the non-stationary setting we will invoke these results with values less than our total time horizon $T$.

\begin{lem}\label{lem:std_lb}
	{\bf \cite[Sec.~5]{Sca17}} For the stationary setting with time horizon $\tau$, if the function $f$ is chosen uniformly at random from the set $\Fc_M$ described above, and if the time horizon satisfies
	\begin{equation} \label{eq:std_lb_le}
		\tau \le O\Big(\frac{M}{\epsilon^2}\Big)
	\end{equation}
	with a small enough implied constant, then the average cumulative regret of any algorithm is lower bounded by $\EE[R_{\tau}]=\Omega(\tau\epsilon)$ provided that $\frac{\epsilon}{B}$ is sufficiently small.  
\end{lem}

Note that in this result, both $M$ and $\epsilon$ may depend on $\tau$.  In \cite{Sca17}, the parameters are chosen such that \eqref{eq:std_lb_le} nearly holds with equality:
\begin{equation} \label{eq:std_lb}
	\tau \asymp \frac{M}{\epsilon^2},
\end{equation}
which combines with \eqref{eq:m_mat_infty} to obtain $\tau \asymp \frac{B^{d/\nu}}{\epsilon^{2+d/\nu}}$ and thus
\begin{equation}  \label{eq:std_eps}
	\epsilon \asymp B^{\frac{d}{2\nu+d}} \tau^{\frac{-\nu}{2\nu+d}}. 
\end{equation}
Combining this with $\EE[R_{\tau}]=\Omega(\tau\epsilon)$  leads to the following lower bound for the Mat\'ern kernel.

\begin{lem}\label{lem:std_main}
	{\bf \cite[Thm.~5]{Sca17}} For the stationary setting with time horizon $\tau$, if the function $f$ is chosen uniformly at random from the set $\Fc_M$ described above, then the cumulative regret of any algorithm is lower bounded by $\EE[R_{\tau}]=\Omega(\tau^{\frac{\nu+d}{2\nu+d}})$ when $B = \Theta(1)$.
\end{lem}

\subsection{Non-Stationary Setting} \label{sec:c_idea}
 
As is typical in lower bounds for non-stationary bandit problems (e.g., see \cite{Bes14} for the finite-arm setting and \cite{cheung2022hedging} for the linear setting), we adopt the high-level strategy of letting the function change after every $\frac{T}{c}$ time steps for some $c \in [1,T]$, so that the time horizon is divided into $c$ ``blocks'' of length $\frac{T}{c}$, and the function stays the same within every block.  (Here we omit rounding as it will not affect our results, but we will later see that care is sometimes needed in ensuring that $c$ stays within the range $[1,T]$.)  In order to utilize Lemmas \ref{lem:std_lb} and \ref{lem:std_main}, we will specifically let the function in each block be chosen uniformly at random from the hard subset $\Fc_M$.

The rough idea is then that each sub-block constitutes a ``stationary sub-problem'', and thus we can lower bound their regret contributions individually and combine them to lower bound the total regret.  While this is conceptually simple, choosing the suitable parameters (e.g., $c$) and ensuring that the relevant constraints (e.g., $\|f_t\| \le B$ and \eqref{eq:restriction}) are satisfied can become a rather delicate process.    The details are given in the following section.

% From now, we focus on noisy BO problem where we have a budget of $\frac{T}{c}$ and $c\in[1,T]$ within each block.  According to the deduction from \cite{Sca17}, we state the following lemma.

\section{Proofs of Main Results}

Having set up the tools from Appendix \ref{sec:tools}, we now proceed to prove the main results, which are given in increasing order of difficulty.  We note that setting $\tau = T$ in Lemma \ref{lem:std_main} immediately gives a lower bound that remains valid in the non-stationary setting (since the stationary setting is a special case that always satisfies \eqref{eq:var1}--\eqref{eq:var3}):
\begin{equation}
	\EE[R_T]  = \Omega\big( T^{\frac{\nu+d}{2\nu+d}} \big). \label{eq:std_main}
\end{equation}
This observation will be useful in suitably limiting the regimes of $\Delta$ that we need to consider.

%\subsection{Lower Bound Tools} \label{sec:tools}
%
%We use the usual bump functions (sometimes modified slightly, but here are the original statements):
%\begin{itemize}
%	\item Height $\epsilon$, width $w$, RKHS norm $B \asymp \frac{\epsilon}{w^{\nu}}$, number of bumps $M = \big(\frac{1}{w}\big)^d$.
%	\item In a length-$\tau$ interval, if $\tau \asymp \frac{M}{\epsilon^2}$ then the regret is order $\tau\epsilon$.
%	\item For the Mat\'ern kernel, $M \asymp \big( \frac{B}{\epsilon} \big)^{d/\nu}$.
%	\item Combining $\tau \asymp \frac{M}{\epsilon^2}$ and $M \asymp \big( \frac{B}{\epsilon} \big)^{d/\nu}$ gives $\epsilon \asymp B^{\frac{d}{2\nu + d}} \tau^{\frac{-\nu}{2\nu + d}}$.
%	\item For the lower bound to be valid, we require $\frac{\epsilon}{B} \to 0$.
%\end{itemize}

\subsection{Proof of Theorem \ref{thm:L} (At Most $L-1$ Changes)}

The case of at most $L-1$ changes (see \eqref{eq:L_changes}) is the most straightforward.  We set $c = L$ in Section \ref{sec:c_idea}, meaning the time horizon $T$ is divided into $L$ intervals of length $T/L$.   %(We could use $c = L+1$, but this would only affect the hidden constant factors.)  
Applying Lemma \ref{lem:std_main} with $\tau = \frac{T}{L}$ for each block, and then summing over the $L$ blocks, we deduce that
\begin{equation}
	\EE[R_T] \gtrsim L\Big(\frac{T}{L}\Big)^{\frac{\nu+d}{2\nu+d}} = L^{\frac{\nu}{2\nu + d}} T^{\frac{\nu+d}{2\nu+d}}
\end{equation}
thus establishing the desired result \eqref{eq:bound_L}.
% Compared to the upper bound, we have $L^{\frac{\nu}{2\nu + d}}$ dependence instead of $\sqrt{L}$, so the two match when $L = O(1)$ or when $\nu$ grows large.  Otherwise there may be a gap, e.g., when $\nu = d$ it's $L^{1/3}$ vs.~$L^{1/2}$.

The statement that $\EE[R_T] \ge \Omega( \sqrt{LT} )$ for all $(\nu,d)$ follows readily from the lower bound \eqref{eq:bound_L}.  This is because the latter is of the form $L^a T^b$ with $a+b=1$, and we have $b = \frac{\nu+d}{2\nu+d} \in \big[\frac{1}{2},1\big]$.  Since $L \le T$, the lower bound is always at least as high as that of $b=\frac{1}{2}$ (and thus $a=\frac{1}{2}$), which is $\sqrt{LT}$.

\subsection{Proof of Theorem \ref{thm:Inf} ($\ell_{\infty}$ Norm Variation)}

We now turn to the case of $\ell_{\infty}$-bounded norm variation, i.e., $\sum_{t=1}^{T-1} \|f_{t+1}-f_{t}\|_{\infty} \le \Delta$.  We start by noting that certain scaling regimes of $\Delta$ can be handled trivially:
\begin{itemize}
	\item If $\Delta = \Theta(T)$, then the regret bound that we seek in \eqref{eq:bound_Inf} is $\Omega(T)$.  Such a result trivially holds because with $\Delta = \Theta(T)$ the function may change arbitrarily (subject to $\|f_t\|_k \le B$) during the first $\Theta(T)$ time steps, recalling that $B$ is a constant.
	\item If $\Delta \lesssim T^{\frac{-\nu}{2\nu + d}}$ then the regret bound that we seek has scaling no higher than $T^{\frac{-\nu^2}{(3\nu+d)(2\nu+d)}} \cdot T^{\frac{2\nu+d}{3\nu+d}} = T^{\frac{-\nu^2 + (2\nu+d)^2}{(3\nu+d)(2\nu+d)}}$.  Expanding the numerator and factorizing gives $-\nu^2 + (2\nu+d)^2 = (3\nu+d)(\nu+d)$, so the scaling simplifies to $T^{\frac{\nu+d}{2\nu+d}}$.  Such a lower bound trivially holds due to the stationary lower bound \eqref{eq:std_main}.
\end{itemize}
Accordingly, it suffices to consider the case that $\Delta \gg T^{\frac{-\nu}{2\nu + d}}$ and $\Delta \ll T$.

We again divide the time horizon into $\frac{T}{c}$ blocks of length $c$, but now the choice of $c$ is dictated by $\Delta$.  Specifically, by \eqref{eq:g_inf} the function changes by at most $4\epsilon$ in $\ell_{\infty}$-norm after each block, and thus we may set $c = \frac{\Delta}{4\epsilon}$ while satisfying the $\Delta$-variation constraint.  The issue of maintaining $c \in [1,T]$ will be addressed below.

Similar to the stationary setting, we choose $\epsilon$ such that \eqref{eq:std_lb} holds, and with $\tau = \frac{T}{c}$ this re-arranges to give 
\begin{equation}
	\epsilon \asymp \Big(\frac{T}{c}\Big)^{\frac{-\nu}{2\nu + d}}. \label{eq:eps_scaling}
\end{equation}
Combining this with $\Delta = 4c\epsilon$ gives $\Delta \asymp c^{\frac{3\nu+d}{2\nu+d}}T^{\frac{-\nu}{2\nu+d}}$, and hence $c \asymp \Delta^{\frac{2\nu+d}{3\nu+d}}T^{\frac{\nu}{3\nu+d}}$.  Note that since we have reduced to the case that $\Delta \gg T^{\frac{-\nu}{2\nu + d}}$ and $\Delta \ll T$, we have $c \gg 1$ and $c \ll T$, thus ensuring the requirement $c \in [1,T]$.

Summing the regret bound from Lemma \ref{lem:std_lb} over the $c$ blocks and then substituting the above choices, we deduce that
\begin{equation}
	\EE[R_T] \gtrsim c \cdot \frac{T}{c} \cdot \epsilon = T\epsilon = \frac{T\Delta}{c} \asymp \Delta^{\frac{\nu}{3\nu+d}}T^{\frac{2\nu+d}{3\nu+d}},
\end{equation}
thus establishing the desired result \eqref{eq:bound_Inf}.

The statement that $\EE[R_T] \ge \Omega( \Delta^{1/3} T^{2/3} )$ for all $(\nu,d)$ follows readily from the lower bound \eqref{eq:bound_Inf}.  This is because the latter is of the form $\Delta^a T^b$ with $a+b=1$, and we have $b = \frac{2\nu+d}{3\nu+d} \in \big[\frac{2}{3},1\big]$.  Since $\Delta \le T$, the lower bound is always at least as high as that of $b = \frac{2}{3}$ (and thus $a=\frac{1}{3}$), which is $\Delta^{1/3} T^{2/3}$.

\subsection{Proof of Theorem \ref{thm:rkhs} (RKHS Norm Variation)}

The proof of Theorem \ref{thm:rkhs} builds on that of Theorem \ref{thm:Inf} but comes with more subtleties and multiple cases to handle.  We again divide the time horizon into $\frac{T}{c}$ blocks of length $c$.  Unlike before, however, we don't necessarily choose the functions to satisfy $\|f_t\|_k = B$.  Instead, we set 
\begin{equation}
	\|f_t\|_k = \min\bigg\{B,\frac{\Delta}{2c}\bigg\}, \label{eq:norm_choice}
\end{equation}
with the second term coming from the variation budget $\Delta$: If each function has RKHS norm $\frac{\Delta}{2c}$, then changing from one function to another incurs a variation cost of at most $\frac{\Delta}{c}$, so after $c$ changes the total cost is at most $\Delta$.

We proceed to study two cases separately based on which term achieves the minimum in \eqref{eq:norm_choice}.

\paragraph{Case 1: $\|f_t\|_k  = B \le \frac{\Delta}{2c}$.}

This case will end end up being used for the case that $\nu \le d$, and accordingly, we seek to establish the regret lower bound stated in \eqref{eq:bound_k1}.  We observe that if $\Delta = O(1)$ then the desired result already follows from \eqref{eq:std_main}, so it suffices to handle $\Delta \gg 1$.  We may also assume $\Delta \ll T$ similar to the proof of Theorem \ref{thm:Inf}.

With $1 \ll \Delta \ll T$ and $\|f_t\|_k = B = \Theta(1)$, we can set $c = \frac{\Delta}{2B}$ while satisfying the variation constraint and maintaining $1 \ll c \ll T$.  We maintain the property from the proof of Theorem \ref{thm:Inf} that $\epsilon \asymp \big(\frac{T}{c}\big)^{\frac{-\nu}{2\nu + d}} = \big(\frac{c}{T}\big)^{\frac{\nu}{2\nu + d}}$ (see \eqref{eq:eps_scaling}), but different from there, we have $\Delta = Bc \asymp c$, and thus $\epsilon \asymp \big( \frac{\Delta}{T}\big)^{\frac{\nu}{2\nu + d}}$.  Note that this satisfies $\epsilon \to 0$ (and thus $\frac{\epsilon}{B} \to 0$ in accordance with the requirement in Lemma \ref{lem:std_lb}) since $\Delta \ll T$.

Summing the regret bound from Lemma \ref{lem:std_lb} over the $c$ blocks and then substituting the above choices, we deduce that
\begin{equation}
	\EE[R_T] \gtrsim c \cdot \frac{T}{c} \cdot \epsilon = T\epsilon \asymp \Delta^{\frac{\nu}{2\nu+d}} T^{\frac{\nu+d}{2\nu+d}}.
\end{equation}
This result is valid for all $(\nu,d)$ provided that $1 \ll \Delta \ll T$, but the next case will provide a better bound for the case that $\nu > d$.

% Since we ended up setting $\Delta \asymp c$ anyway (and thus the two terms in $\min\big\{B,\frac{\Delta}{2c}\big\}$ coincide to within a constant factor), it may be that this case isn't needed, and what we need might already be fully captured by Case 2.

\paragraph{Case 2: $\|f_t\|_k = \frac{\Delta}{2c} \le B$.}

In this case, when applying the results in Appendix \ref{sec:tools}, we need to take care in the fact that $B$ therein represents $\|f_t\|_k$, and we should thus replace it by  $\frac{\Delta}{2c}$ everywhere.  This gives the following:
\begin{itemize}
	\item The scaling of $\epsilon$ in \eqref{eq:std_eps} becomes $\epsilon \asymp \big( \frac{\Delta}{c} \big)^{\frac{d}{2\nu + d}} \tau^{\frac{-\nu}{2\nu + d}}$.
	\item The requirement ``$\frac{\epsilon}{B}$ is sufficiently small'' in Lemma \ref{lem:std_lb} is replaced by ``$\frac{\epsilon c}{\Delta}$ is sufficiently small''.
\end{itemize}
The former of these, along with $\tau = \frac{T}{c}$, gives the regret bound
\begin{equation}
	\EE[R_T] \gtrsim c \cdot \frac{T}{c} \cdot \epsilon = T\epsilon \asymp  \Delta^{\frac{d}{2\nu + d}} T^{\frac{\nu+d}{2\nu + d}} \Big( \frac{1}{c} \Big)^{\frac{d-\nu}{2\nu + d}}. \label{eq:RT_Case2}
\end{equation}
This suggests that to get the strongest lower bound, we should try to make $c$ as small as possible if $d > \nu$, but as large as possible if $d < \nu$.  However, we need to be careful to satisfy above-established condition $\min\big\{B,\frac{\Delta}{2c}\big\} = \frac{\Delta}{2c}$ and  $\frac{\epsilon c}{\Delta}$ being sufficiently small, as well as $c \in [1,T]$.  We proceed formally as follows.

{\bf Sub-case (a) ($d \ge \nu$).} Having $\min\big\{B,\frac{\Delta}{2c}\big\} = \frac{\Delta}{2c}$ requires $c \gtrsim \frac{\Delta}{B} \asymp \Delta$, so to make $c$ as small as possible, we choose $c \asymp \Delta$.  Thus, we get the same regret bound as Case 1, matching \eqref{eq:bound_k1}:
\begin{equation}
	\EE[R_T] \gtrsim T\epsilon \asymp  \Delta^{\frac{\nu}{2\nu + d}} T^{\frac{\nu+d}{2\nu + d}}.
\end{equation}
We also have $\epsilon \to 0$ and $c \in [1,T]$ (resulting from $1 \ll \Delta \ll T$) due to the reasoning given in Case 1.

%In Case 1 we assumed that $\Delta \gg 1$, which may seem restrictive.  But if $\Delta \ll 1$ then we might as well have no changes and just use the standard $T^{\frac{\nu+d}{2\nu + d}}$ lower bound, since any multiplication by $\Delta^{\rm \alpha}$ (with $\alpha > 0$) will only make the bound worse.  In other words, we could write the lower bound as
%\begin{equation}
%	\EE[R_T] \gtrsim \max\Big\{1, \Delta^{\frac{\nu}{2\nu + d}} \Big\} T^{\frac{\nu+d}{2\nu + d}}.
%\end{equation}
%This suggests that there might be no performance degradation when $\Delta = O(1)$, unlike the other cases.  This could be a weakness in the analysis, or it could be that \emph{small-$\nu$ settings, while incurring more regret, are relatively less impacted by small time variations}.

{\bf Sub-case (b) ($\nu \ge d$).} In this case, the above-established condition
\begin{equation}
	\frac{\epsilon c}{\Delta} \text{ is sufficiently small}
\end{equation}
prevents us from letting $c$ be too large.  To understand this condition, we note that the above-mentioned scaling $\epsilon \asymp \big( \frac{\Delta}{c} \big)^{\frac{d}{2\nu + d}} \big(\frac{T}{c} \big)^{\frac{-\nu}{2\nu + d}}$ gives $\frac{\epsilon c}{\Delta} \asymp \big( \frac{c}{\Delta} \big)^{\frac{2\nu}{2\nu + d}} \big(\frac{T}{c} \big)^{\frac{-\nu}{2\nu + d}}$, which equals 1 when $\big(\frac{c}{\Delta}\big)^2 = \frac{T}{c} \iff c = \big(T \Delta^2 \big)^{1/3}$.  Accordingly, in order for $\frac{\epsilon c}{\Delta}$ to be sufficiently small (as required above), it suffices that $c \lesssim \big(T \Delta^2 \big)^{1/3}$ with a sufficiently small implied constant.  We let this hold roughly with equality:
\begin{equation}
	c \asymp \big(T \Delta^2 \big)^{1/3} \label{eq:c_choice}
\end{equation}
with a small enough implied constant.  (The condition $c \in [1,T]$ is discussed below.)  Thus,
\begin{equation}
	\bigg( \frac{1}{c} \bigg)^{\frac{d-\nu}{2\nu+d}} \asymp \big( T \Delta^2 \big)^{ \frac{\frac{1}{3}(\nu - d)}{2\nu + d}},
\end{equation}
and substituting into \eqref{eq:RT_Case2} gives
\begin{equation}
	\EE[R_T] \gtrsim T\epsilon \asymp  \Delta^{\frac{2\nu/3+d/3}{2\nu + d}} T^{\frac{4\nu/3+2d/3}{2\nu + d}} = \Delta^{1/3}T^{2/3}.
\end{equation}
% Note that compared to the upper bound in \eqref{eq:RT_Matern}, the exponent is $\frac{4\nu/3+2d/3}{2\nu+d}$ instead of $\frac{4\nu/3 + d}{2\nu+d}$, so the two differ by a factor of $T^{\frac{d/3}{2\nu+d}}$.

It remains to verify that \eqref{eq:c_choice} satisfies $c \in [1,T]$.  We have already established that it suffices to consider $\Delta \ll T$, so the condition $c \le T$ is immediate.  On the other hand, we observe that $c \gg 1$ if only if $\Delta \gg \frac{1}{\sqrt T}$.  To see that we can restrict our attention to such a scenario, we note that if $\Delta \lesssim \frac{1}{\sqrt T}$ then the scaling $\Delta^{1/3}T^{2/3}$ behaves as $o(\sqrt T)$, which means that there is nothing to prove because the stationary lower bound in \eqref{eq:std_lb} is already $\Omega( \sqrt{T} )$ (or higher).  Thus, we have established the desired result \eqref{eq:bound_k2}.

\paragraph{Last part of the theorem.} The statement that $\EE[R_T] \ge \Omega( \Delta^{1/3} T^{2/3} )$ for all $(\nu,d)$ follows readily from the lower bounds \eqref{eq:bound_k1}--\eqref{eq:bound_k2}.  This is seen as follows:
\begin{itemize}
	\item For $\nu \le d$, we see that \eqref{eq:bound_k1} is of the form $\Delta^a T^b$ with $a+b=1$, and we have $b = \frac{\nu+d}{2\nu+d}$, which lies in $\big[\frac{2}{3},1\big]$ when $\nu \le d$. Since $\Delta \le T$, the lower bound is always at least as high as that of $b = \frac{2}{3}$ (and thus $a=\frac{1}{3}$), which is $\Delta^{1/3} T^{2/3}$.
	\item For $\nu \ge d$, we see that \eqref{eq:bound_k2} already scales as $\Delta^{1/3} T^{2/3}$.
\end{itemize}

\rev{
\subsection{Discussion on the Squared Exponential Kernel} \label{app:SE}

The squared exponential (SE) kernel (normalized to satisfy $k(x,x') = 1$) is given as follows:
\begin{equation}
    \kSE(x,x') = \exp \bigg(- \dfrac{\|x - x'\|^2}{2\ell^2} \bigg), \label{eq:kSE}
\end{equation}
where $\ell > 0$ is the lengthscale.  It is widely understood that this kernel behaves similarly to the Mat\'ern in the limit of its smoothness parameter $\nu$ approaching $\infty$, and this is refelcted in existing regret bounds (e.g., cumulative regret with dependence $T^{\frac{\nu+d}{\nu+2d}}$ in the standard setting approaching $T^{\frac{1}{2}}$). 

In our setting, the preceding proofs can readily be adapted to the SE kernel by using \eqref{eq:m_se_infty} in place of \eqref{eq:m_mat_infty}.  The subsequent amount of effort required depends on how loose one is willing to be with logarithmic factors.  If one is willing to ignore $(\log T)^{O(d)}$ factors, then we can readily replace \eqref{eq:std_eps} by $\epsilon \stackrel{\log}{\asymp} \tau^{-\frac{1}{2}}$ where the notation $\stackrel{\log}{\asymp}$ ignores logarithmic factors, and then proceed similarly to the preceding subsections to obtain regret bounds of $\widetilde{\Omega}( \sqrt{LT} )$ ($L$ switches) and $\widetilde{\Omega}\big( \Delta^{1/3} T^{2/3} \big)$ ($\ell_{\infty}$ and RKHS norm variation), with $\widetilde{\Omega}(\cdot)$ again hiding logarithmic factors.  

On the other hand, more care is needed when keeping track of the logarithmic factors, as was done for lower bounds in the standard setting \cite{Sca17} (in order to get a lower bound of $\Omega(\sqrt{T(\log T)^{d/2}})$ which is more precise than $\widetilde{\Omega}(\sqrt{T})$).  We can readily do the same here, but we omit the details because our central focus is the Mat\'ern kernel.
}

\section{MASTER Reduction of GP-UCB}\label{app:master}

In this section, we primarily rely on the key results derived by \cite[App.~E]{Hon23}, as well as the MASTER reduction proposed by \cite{Wei21}.  
Throughout the section, we let $\Delta_{1:t} = \sum_{i=1}^{t} \|f_{i+1}-f_{i}\|_{\infty}$ denote the $\ell_{\infty}$-norm variation up to time $t$, and we let $f_i^*$ be the maximal value of function $f_i$.

\subsection{General MASTER Reduction Result} \label{sec:master_gen}

The following lemma states a variation of the conditions for the MASTER reduction method to be applicable.

\begin{assump}\label{lem:master_cond}
    {\bf \cite[Cond.~E.1]{Hon23}}  Assume that there exists a value $\zeta$ and a sequence $\{\rho_t\}_{t=1}^T$ such that $\zeta \Delta_{1:t} \le \rho_t$ for all $t \in \{1,\dotsc,T\}$, and assume that a base algorithm can produce $\ftil_t$, using the history up to $t-1$, such that the following conditions are satisfied:
    \begin{equation}
        \ftil_t\ge \min_{i=1,\dotsc,t} f_i^* - \zeta \Delta_{1:t}, \quad \text{ and } \frac{1}{t} \sum_{\tau=1}^{t}(\ftil_t - y_\tau) \le c_1\rho_t + c_1\zeta \Delta_{1:t},
    \end{equation}
    with probability at least $1-\frac{\delta}{T}$, where $c_1$ is a universal constant.  Furthermore, assume that $\rho_t \ge \frac{1}{\sqrt{t}}$, and that $t\rho_t$ is non-decreasing in $t$.
\end{assump}

Under this assumption, we have the following, which is a variation of \cite[Thm.~2]{Wei21}.

\begin{thm}\label{thm:master}
	{\bf \cite[Thm.~E.2]{Hon23}} If a base algorithm satisfies Assumption \ref{lem:master_cond} with $t\rho_t= g_{1,t}\sqrt{t} + g_{2,t}$ for non-decreasing $\{g_{1,t}\}_{t=1}^T$ and $\{g_{2,t}\}_{t=1}^T$,  then the algorithm obtained via the MASTER reduction guarantees that
	\begin{equation}\label{eq:master}
		\EE[R_T] = \Otil\Big( \big(g_{1,T}^{\frac{2}{3}} + g_{2,T}g_{1,T}^{-\frac{4}{3}}\big)\zeta^{\frac{1}{3}}\Delta^{\frac{1}{3}}T^{\frac{2}{3}} + \big(g_{1,T} + g_{1,T}^{-1} g_{2,T}\big)\sqrt{T} \Big).
	\end{equation}
\end{thm}

\subsection{Existing Application to the Kernelized Setting} \label{sec:master_old}

Before discussing the application of Theorem \ref{thm:master} to our setting, we introduce some standard tools from the kernelized bandit literature.  Given a dataset $\{(\xv_t, y_t)\}_{t=1}^{n}$ with cardinality $n$, it is natural to consider a ``fictitious'' Gaussian process prior $\GP(0,k)$, and approximate the black-box function $f$ through the following GP posterior with regularization parameter $\lambda > 0$:
\begin{align}
	\mu_{n}(\xv) &= \kv_n(\xv)^T\big(\Kv_n + \lambda \mathbf{I}_n \big)^{-1} \yv_n,  \label{eq:posterior_mean} \\ 
	\sigma_{n}^2(\xv) &= k(\xv,\xv) - \kv_n(\xv)^T \big(\Kv_n + \lambda \mathbf{I}_n \big)^{-1} \kv_n(\xv),\label{eq:posterior_variance}
\end{align}
where $\yv_n = [y_1,\ldots,y_n]^T$, $\kv_n(\xv) = \big[k(\xv_t,\xv)\big]_{t=1}^n$, $\Kv_n = \big[k(\xv_t,\xv_{t'})\big]_{t,t'}$ is the kernel matrix, $\mathbf{I}_n$ is the identity matrix of dimension $n$.  We also recall the well-known definition of \emph{maximum information gain} \cite{Sri09,Vak21a}:
\begin{equation} 
	\gamma_T = \max_{x_1,\dotsc,x_T} \max_{S \,:\, |S| = T} I(f;\mathbf{y}_S), \qquad f \sim \mathcal{GP}(0,k), \label{eq:gamma_def}
\end{equation}
with $I(X;Y)$ denoting mutual information.   The quantity $\gamma_T$ represents the maximum amount of information that a set of $T$ observations can reveal about a zero-mean Gaussian process $f$ with kernel $k$, and despite its Bayesian definition, it also plays a fundamental role in the frequentist setting with RKHS functions.

In the stationary setting in which $f_t = f, \forall t$, the use of equations \eqref{eq:posterior_mean} and \eqref{eq:posterior_variance} are well-justified by \emph{confidence bounds} of the following form:
\begin{align}
	{\rm UCB}_t(\xv) &= \mu_{t-1}(\xv) + \beta_t \sigma_{t-1}(\xv) \\
	{\rm LCB}_t(\xv) &= \mu_{t-1}(\xv) - \beta_t \sigma_{t-1}(\xv).
\end{align}
Specifically, under a suitable choice of $\beta_t$ (e.g., see Lemma \ref{lem:gpucb_master_cond} below), we are guaranteed that the following holds with high probability \cite{Sri09,Cho17}:
\begin{equation}
	{\rm LCB}_t(\xv) \le f(\xv) \le {\rm UCB}_t(\xv), ~~ \forall \xv,t.
\end{equation}
In the non-stationary setting, we cannot directly make such a statement (with $f_t$ replacing $f$), but we can still make use of the confidence bounds via the MASTER reduction.

In \cite[App.~E]{Hon23}, Theorem \ref{thm:master} was applied to GP-UCB with the most well-known confidence bounds (e.g., see \cite{Sri09,Cho17}), dictated by the choice of $\beta_t$ in the following lemma statement.  % We recall that $\gamma_T$ denotes the maximum information gain associated with the kernel (e.g., see \cite{Sri09,Vak21a}).

\begin{lem}\label{lem:gpucb_master_cond}
	{\bf \cite[Lem.~E.3]{Hon23}} The GP-UCB algorithm satisfies Lemma \ref{lem:master_cond} with $\ftil_t = \mu_{t-1}(\xv) + \beta_t \sigma_{t-1}(\xv)$, with $\rho_t = \beta_t \sqrt{\frac{\gamma_t\log(T/\delta)}{{t}}}$, $\zeta = \gamma_T \sqrt{\log(T/\delta)}$, and $\beta_t = \sqrt{\lambda}B+\sigma\sqrt{2\gamma_t+2\log(1/\delta)}$.  % Here $\mu_t(\cdot)$ and $\sigma_t(\cdot)$ are the stationary posterior mean and standard deviations under a Gaussian process prior with regularization parameter $\lambda > 0$, and this parameter $\lambda$ may be chosen arbitrarily.  
\end{lem}

\subsection{Limitations of the MASTER Reduction in the Kernelized Setting} \label{sec:master_new}

In \cite{Hon23}, Lemma \ref{lem:gpucb_master_cond} was shown to achieve a worse regret bound than the optimization-based approach in \cite{Hon23}; the two scalings are $\Otil( \gamma_T^{1/3} \Delta^{1/3} T^{2/3})$ vs.~$\Otil( \gamma_T \Delta^{1/3} T^{2/3})$.  However, it is well-known that existing GP-UCB analyses are suboptimal even in the stationary setting \cite{vakili2021open}, and this raises the interesting question of whether future improved confidence bounds may lead to a better MASTER reduction.

One approach to addressing this question would be to study improved GP-UCB type algorithms that have recently been proposed, e.g., see \cite{Whi23}.  However, to understand both existing and \emph{potential future improvements}, we instead consider the \emph{hypothetical scenario where $\beta_t = 1$}.  This is justified by the fact that if we were to have $\beta_t \ll 1$, then following the stationary GP-UCB analysis would readily lead to an upper bound with $T$-dependence strictly better than $\sqrt{T \gamma_T}$, which would contradict the lower bounds from \cite{Sca17,Cai21}.  Thus, by considering $\beta_t = 1$, we are essentially studying the \emph{best-case scenario} of what the MASTER reduction might provide (at least using its current analysis tools) if the confidence bounds for RKHS functions were to improve in the future.

Observe that if Lemma \ref{lem:gpucb_master_cond}  were to hold with $\beta_t = 1$, we would have $t\rho_t = \sqrt{t}\sqrt{\gamma_t\log(t/\delta)}$.  Thus, $g_{1,t}$ and $g_{2,t}$ in Theorem \ref{thm:master} would be given by $g_{1,t} =  \sqrt{\gamma_t\log(T/\delta)}$ and $g_{2,t}=0$.  Substituting these into \eqref{eq:master}, we would obtain the following \emph{hypothetical} upper bound:
\begin{align}
    \EE[R_T] &= \Otil\Big( g_{1,T}^{\frac{2}{3}} \zeta^{\frac{1}{3}}\Delta^{\frac{1}{3}}T^{\frac{2}{3}} + g_{1,T}\sqrt{T} \Big) \label{eq:master-ucb1}\\
    &= \Otil\Big( \gamma_T^{\frac{1}{3}} \gamma_T^{\frac{1}{3}} \Delta^{\frac{1}{3}}T^{\frac{2}{3}} + \sqrt{T\gamma_T} \Big) \label{eq:master-ucb2}\\
    &= \Otil\Big(\gamma_T^{\frac{2}{3}} \Delta^{\frac{1}{3}}T^{\frac{2}{3}} + \sqrt{T\gamma_T} \Big). \label{eq:master-ucb3}
\end{align}
We observe that the $\Otil( \gamma_T^{2/3} \Delta^{1/3} T^{2/3})$ scaling is still worse than the $\Otil( \gamma_T^{1/3} \Delta^{1/3} T^{2/3})$ scaling of \cite{Hon23}.  Thus, either the MASTER reduction approach is fundamentally limited in the kernelized setting, or the current tools for analyzing it have room for improvement.

\end{document}